\documentclass[10pt,twocolumn,letterpaper]{article}

\usepackage{cvpr}
\usepackage{times}
\usepackage{epsfig}
\usepackage{graphicx}
\usepackage{amsmath}
\usepackage{amssymb}

\usepackage{lipsum}
\usepackage[ruled,vlined]{algorithm2e}
\usepackage{subcaption}

\usepackage[pagebackref=true,breaklinks=true,letterpaper=true,colorlinks,bookmarks=false]{hyperref}

\cvprfinalcopy 

\def\bd{\mathbf d}

\def\TK{\mathbf K}

\def\bp{\mathbf p}
\def\by{\mathbf y}

\def\bp{\mathbf p}

\def\bc{\mathbf c}

\def\TQ{\mathbf Q}

\newcommand{\grad}{\ensuremath{\nabla}}

\ifcvprfinal\fi
\begin{document}

\title{Deep connections between learning from limited labels \& physical parameter estimation - inspiration for regularization}

\author{Bas Peters\\
Computational Geosciences Inc.\\
1623 West 2nd Ave, Vancouver, BC, Canada\\
{\tt\small bas@compgeoinc.com}
}

\maketitle

\begin{abstract}
Recently established equivalences between differential equations and the structure of neural networks enabled some interpretation of training of a neural network as partial-differential-equation (PDE) constrained optimization. We add to the previously established connections, explicit regularization that is particularly beneficial in the case of single large-scale examples with partial annotation. We show that explicit regularization of model parameters in PDE constrained optimization translates to regularization of the network output. Examination of the structure of the corresponding Lagrangian and backpropagation algorithm do not reveal additional computational challenges. A hyperspectral imaging example shows that minimum prior information together with cross-validation for optimal regularization parameters boosts the segmentation accuracy.
\end{abstract}

\section{Introduction}
Although many well-known computer vision benchmarks come with hundreds or more fully annotated images, many real-world applications come with few labels, and it is impossible/very costly to collect more labels/ground-truth.

Working with limited labels/annotation is also the default for many inverse problems. For instance, partial-differential-equation (PDE) constrained optimization for parameter estimation aims to estimate physical model parameters that predict the observations (labels). Examples include acoustic velocity estimation from observed seismic waves or conductivity from electromagnetic fields. Most inverse problems cannot be `solved' using just the partially observed fields and a physical model that connects input and output. Prior knowledge about the underlying mathematical structure of the quantity to be estimated is typically necessary, i.e., regularization.

In this work, we focus on PDE-constrained optimization problems and regularization, describe their deep connection to learning from limited labels using neural networks, and show that there are subtle differences between the two tasks. Despite these differences, we show how to transfer regularization ideas from PDE-constrained parameter estimation to help training neural networks in the case of limited labels.

Recently, researchers established a connection between time-stepping methods for solving differential equations and deep neural networks \cite{HaberRuthotto2017a,RuthottoHaber2018,neuroODE}. This includes the recognition of the standard ResNet \cite{he2016deep} as a time-dependent heat equation, deep neural networks based on the reaction-diffusion (advection) equation \cite{pmlr-v97-haber19a,ANODEV2}, as well as second-order equations like the nonlinear telegraph equation \cite{Chang2017Reversible}. These contributions concern various discretizations of the forward problem in the context of inverse problems. 

Regarding optimization, \cite{LeCunBackPropLagrangian} showed that the backpropagation algorithm is equivalent to an implementation of the method of Lagrangian multipliers for equality constrained nonlinear optimization. Stating the training of a network as constrained optimization opens the door to methods other than backpropagation: for instance, \cite{pmlr-v48-taylor16,doi:10.1137/19M1247620} increase parallelism and \cite{ijcai2019-103,lensink2019fully,peters2019symmetric} exploit the connection to PDE-constrained optimization to reduce memory. In this work, we
\begin{itemize}
\item show that regularization on the model parameters in PDE-constrained optimization translates to regularization of the output of a neural network;
\item derive the Lagrangian and backpropagation algorithm for this type of regularization, which reveals no additional computational complications;
\item network-output regularization based on minimal prior knowledge can boost the prediction accuracy when training on few labels.
\end{itemize}

First, we establish the correspondence between the terminology in deep learning and PDE-constrained optimization. Next, we state the problems and derive an optimization algorithm for both regularized PDE-constrained optimization and training deep neural networks. An example shows that basic assumptions about the output of a network for a given application lead to a simple, yet effective method to improve semantic segmentation results in the label-sparse training regime.

\section{Terminology differences}

In PDE-constrained optimization, the \emph{model} is a physical differential equation, such as the heat, or Maxwell equations.  The \emph{model parameters} are the coefficients of the equation, i.e., the spatial distribution of acoustic velocity or electrical conductivity. The input to the model are the initial conditions, or source functions. The output of the physical system is compared to the observed \emph{data}.
When training a neural network, we commonly refer to the input of a network as the \emph{data}. The output (prediction) of the network is compared to the \emph{labels}. The \emph{model} is the structure of the neural network; examples of  \emph{model parameters} are biasses and convolutional kernels. The goal of image segmentation is to estimate model parameters that transform the input data (images) into the labels (segmentation). 

\section{Output-regularized neural network training as PDE-constrained optimization}
Rather than standard tensor notation, we employ matrix-vector product descriptions to stay close to PDE-constrained optimization literature. We block-vectorize states and Lagrangian multipliers, while weight/parameter tensors are flattened into block-matrices, see \cite{RuthottoHaber2018,doi:10.1190/segam2019-3216640.1,LeanConv8989808}. To keep notation compact, we focus the ResNet \cite{he2016deep} with time-step $h$. The network state $\by_j$ at layer $j$ is given by
\begin{equation}
\by_j = \by_{j-1} - h f(\TK_j \by_{j-1}).
\end{equation}
The block-matrix $\TK$ denotes the network parameters where the number of block-rows is equal to the number of output channels, the number of block-columns is equal to the number of input channels. The nonlinear pointwise $f$ is the activation function. Given data $\bd$ and labels $\bc$, we want to minimize a loss function $l(\bd,\{\TK\},\bc)$, subject to one equality constraint per time-step, i.e.,

\begin{align}\label{prob}
\min_{\{\TK\}} \:\: &l(\TQ \by_n  ,  \bc) + \alpha  R(\by_n) \:\: \text{s.t.} \\
&\by_n = \by_{n-1} - f(\TK_n \by_{n-1}) \nonumber \\
&\vdots \nonumber \\
&\by_j = \by_{j-1} - f(\TK_j \by_{j-1}) \nonumber  \\
&\vdots \nonumber \\
&\by_1 = \bd. \nonumber
\end{align}
There are $n$ time-steps (network layers). The matrix $\TQ$ selects from the prediction, the pixel indices where we have labels. This is analogous to restricting physical fields to sensor locations \cite{doi:10.1190/INT-2018-0225.1,doi:10.1190/tle38070534.1}. Compared to earlier regularized PDE-constrained optimization formulations \cite{HaberRuthotto2017a}, we propose to apply $\alpha  R(\by_n)$ to the network output (and not on the parameters $\TK$ as in weight-decay, or parameter smoothness in time  \cite{HaberRuthotto2017a}), such that we can explicitly control the properties of the predicted probability maps. This is similar in spirit to \cite{topoawarenetworks}, see also \cite{kukavcka2017regularization} for context regarding other types of implicit/explicit regularization. While we could employ automatic differentiation to the above problem, we need to look at the Lagrangian to see what are the effects of output regularization on the subsequent optimization procedures:
\begin{align}
&L(\{\by\},\{\bp\},\{\TK\}) = l(\TQ \by_n  , \bd) + \alpha  R(\by_n) \\
- &\sum_{j=2}^n  \bp_j^\top (\by_j - \by_{j-1} + f(\TK_j \by_{j-1})) -  \bp_1^\top (\by_1 - \bd) \nonumber ,
\end{align}
where $\bp_j$ are the Lagrangian multipliers. To construct an algorithm for problem \eqref{prob}, we need the partial derivatives of $L$ w.r.t. the variables,
\begin{align}\label{LagGrads}
&\grad_{\by_n} L = \grad_{\by_n} l(\TQ \by_n  , \bd) + \alpha  \grad_{\by_n} R(\by_n) - \bp_n\\
&\nonumber\\
&\text{for} \: j=n-1,\cdots,2 : \nonumber\\
&\grad_{\by_{j-1}} L  = -\bp_{j-1} + \bp_j -\TK_{j-1}^\top  \operatorname{diag}(f'(\TK_{j-1} \by_{j-1}))  \bp_{j} \\
&\grad_{\bp_j} L = -\by_j + \by_{j-1} - f(\TK_j \by_{j-1})\\
&\grad_{\TK_j} L = \bigg[ \frac{\partial \big(\ \TK_j \by_{j-1} \big)}{\partial \TK_j}\bigg]^\top \operatorname{diag}(f'(\TK_j \by_{j-1})) \bp_j
\end{align}
where we left out the derivatives related to the first layer as they are the initial condition that we set equal to the input data while training. $f'$ denotes the derivative of $f$. The above shows that the derivatives of the regularization term appear in the partial derivative of the Lagrangrian with respect to the last state ($\by_n$) only. To compute a gradient, we use the `standard' tools: adjoint-state/backpropagation, see \cite{LeCunBackPropLagrangian,ijcai2019-103} for details about their equivalence. Note that to satisfy the first-order optimality conditions for \eqref{prob}, each of the partial derivatives in \eqref{LagGrads} needs to vanish: 1) Forward propagate through the network to satisfy all constraints in \eqref{prob}; $\grad_{\bp_j} L$ vanishes. 2) Propagate backward to obtain all Lagrangian multipliers $\bp_j$. 3) compute gradients w.r.t. the network parameters $\TK_j$ for every layer. In Algorithm \ref{alg:backprop} we show a slightly different version that computes the gradient w.r.t. parameters on the fly. This procedure avoids storing more Lagrangian multipliers than the length of the recursion of the differential equation discretization, instead of storing the multipliers for all layers. Note that it is possible to avoid storing all states $\by_j$ via reversible networks that re-compute the states on the fly during backpropagation \cite{lensink2019fully}. 

\begin{algorithm}[]
\SetAlgoLined
 $\by_{1} = \bd$  //Initialization \; 
 \For{$i=2,\cdots,n$}{
  $\by_j = \by_{j-1} - f(\TK_j \by_{j-1}) $ // Forward\;
 }
Compute final Lagrangian multiplier;
 $\bp_{n} = \grad_{\by_n} l(\TQ \by_n  , \bd) + \alpha  \grad_{\by_n} R(\by_n)$ \;
//Propagate backward\;
 \For{$i=n,n-1,\cdots,2$}{
 $\grad_{\TK_j} L = \bigg[ \frac{\partial \big(\ \TK_j \by_{j-1} \big)}{\partial \TK_j}\bigg]^\top \operatorname{diag}(f'(\TK_j \by_{j-1}) \bp_j$ \;
  $ \bp_{j-1} = \bp_j -\TK_{j-1}^\top  \operatorname{diag}(f'(\TK_{j-1} \by_{j-1})) \bp_{j}$\;
}
 \caption{adjoint-state/backpropagation to compute the gradient of a network}
\label{alg:backprop}
\end{algorithm}

\section{Examples of simple explicit regularizers}
So far we showed that PDE-constrained optimization and training neural-networks are similar processes. However, the goals are different. PDE-constrained optimization often estimates material properties (parameters) with the help of an additive scaled regularization function $R(\TK)$. In contrast, when training networks, we primarily care about the prediction (network output). Despite these different objectives, a successful technique to prevent overfitting is regularization of the network parameters, just as in the PDE-constrained optimization case. This is a form of implicit regularization, as it is not trivial to see how the regularization relates to the visual appearance of the output for applications like semantic segmentation and other image-to-image applications. In the PDE-constrained optimization setting, regularizing the parameters is a form of explicit regularization because the regularization directly acts on the quantity of interest.

The second contribution of this work is to recognize that we can carry over the explicit regularization nature from PDE-constrained optimization to training neural networks by adding a regularization (penalty) term on the network output (final state $\by_n$).

Consider semantic segmentation in the case of limited data and even more limited supervision. Specifically, consider applications with a single (possibly large-scale) example with partial labeling or point annotations \cite{doi:10.1190/INT-2018-0225.1} where it is difficult or impossible to collect additional examples. In the next section, we present results for such an application: time-lapse hyperspectral imaging using partial labeling. Insufficient annotation leads to high-frequency/oscillatory artifacts in the final network state, or, probability map, $\by_n$. Using the prior knowledge that the true thresholded prediction is piecewise-constant and there are not many isolated pixels or line segments with a class different from their surroundings, a reasonable choice for regularization is a quadratic smoother
\begin{equation}\label{quadsmt}
R(\by_n) = \frac{1}{2}\|\nabla_1 \by_n \|_2^2 + \frac{1}{2}\| \nabla_2 \by_n \|_2^2
\end{equation}
with gradient
\begin{equation}\label{quadsmt_grad}
\nabla_{\by_n} R(\by_n) = \nabla_1^\top \nabla_1 \by_n + \nabla_2^\top \nabla_2 \by_n.
\end{equation}
The discrete gradients $\nabla_1$ and $\nabla_2$ are the derivatives of an image in the first or second coordinate, respectively. This regularizer adds to the final Lagrangian multiplier. Subsequently, this information backpropagates and influences the gradient w.r.t. the network parameters $\nabla_{\TK}L$. In the above example, the quadratic smoothing makes sure the final output state transitions across class boundaries smoothly.

\section{Choice of $\alpha$ and numerical example}
\begin{figure}[!htb]
 	\centering
 	\begin{subfigure}[b]{0.23\textwidth}
 		\includegraphics[width=\textwidth]{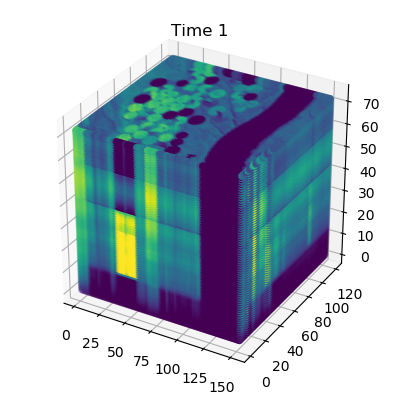}
 		\caption{}
 		\label{fig:Figure1a}
 	\end{subfigure}
 	\begin{subfigure}[b]{0.23\textwidth}
 		\includegraphics[width=\textwidth]{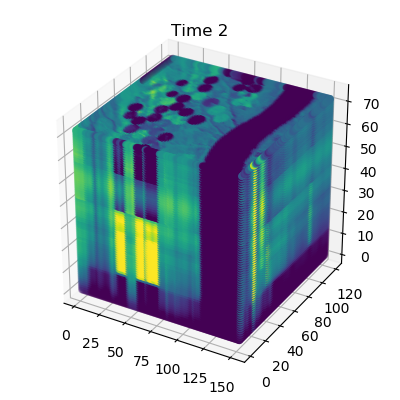}
 		\caption{}
 		\label{fig:Figure1b}
 	\end{subfigure}
 	\begin{subfigure}[b]{0.2\textwidth}
 		\includegraphics[width=\textwidth]{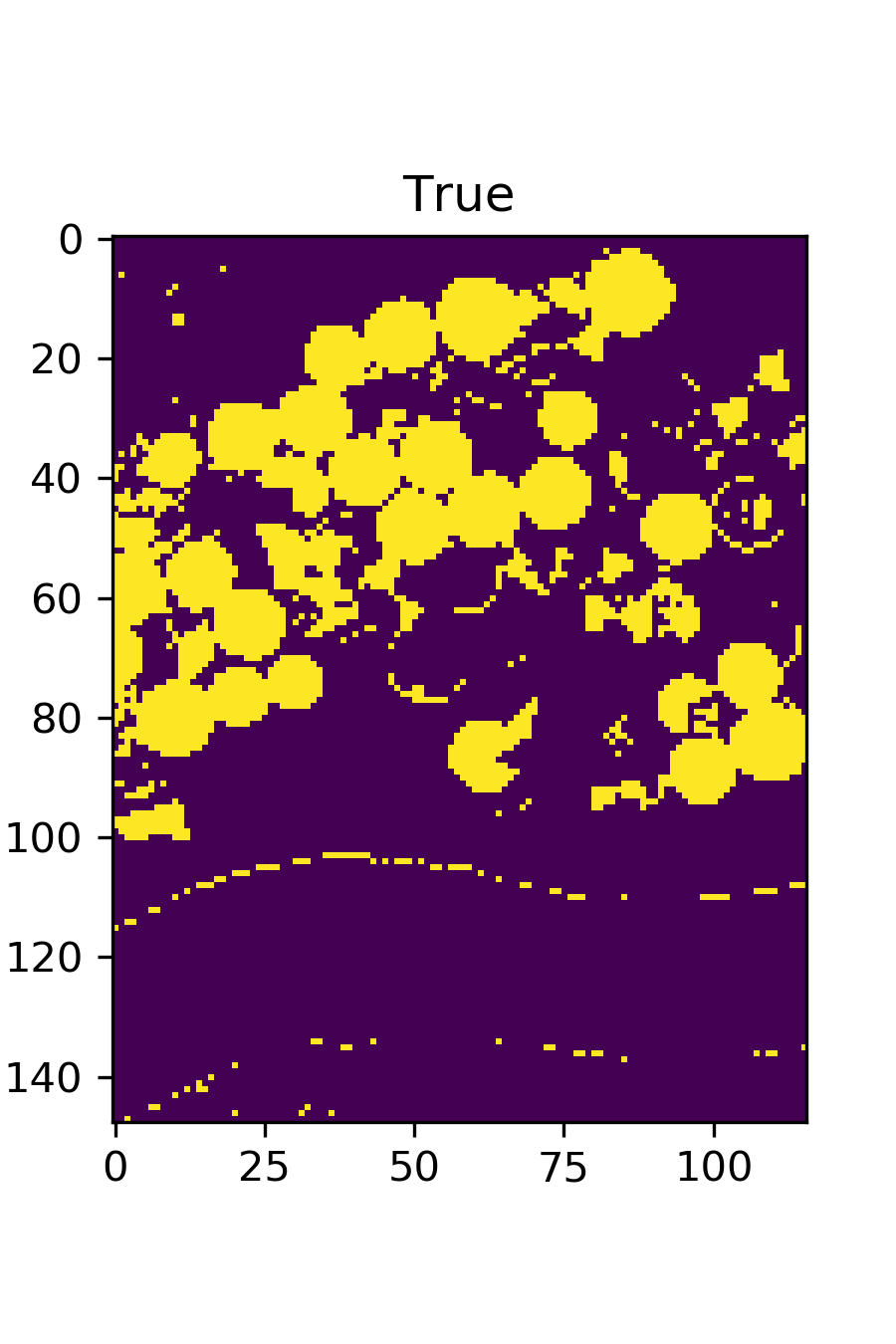}
 		\caption{}
 		\label{fig:Figure1c}
 	\end{subfigure}
 	\caption{(a) and (b): Two time-instances of hyperspectral data collected over the same location. (c) true land-use change (a 2D map of the surface of the Earth).}
\label{fig:data}
 \end{figure}

This example shows that 1) while the overall optimization problem is still non-convex, cross-validation to select the regularization parameter $\alpha$ results in expected behavior; 2)  the proposed explicit regularization can improve predictions. The task is land-use change detection from time-lapse (4D) hyperspectral data. The input data \cite{doi:10.1080/01431161.2018.1466079} are two 3D hyperspectral datasets collected over the same location, with two spatial and one frequency axis, see Figure \ref{fig:data}. 

The goal is to output a 2D map of the earth's surface that shows land-use change (Figure \ref{fig:data}). For training, there are $200$ randomly selected annotated pixels; and $50$ pixels for validation. The available labels were annotated by an expert. The default mode of operation for many hyperspectral tasks is interpolation/extrapolation of the labels to the full map, see, e.g., \cite{C7729859,M8297014,doi:10.1080/2150704X.2019.1681598,rs9010067,rs12010188}. Generalization to new locations is not a concern.  The difficulty of annotation causes limited availability of annotated hyperspectral data; the person that creates the labels needs to know the application and nature of hyperspectral data. Alternatively, the labels come from costly ground truth observations.

   \begin{figure}[!htb]
   \centering
   \includegraphics[width=0.4\textwidth]{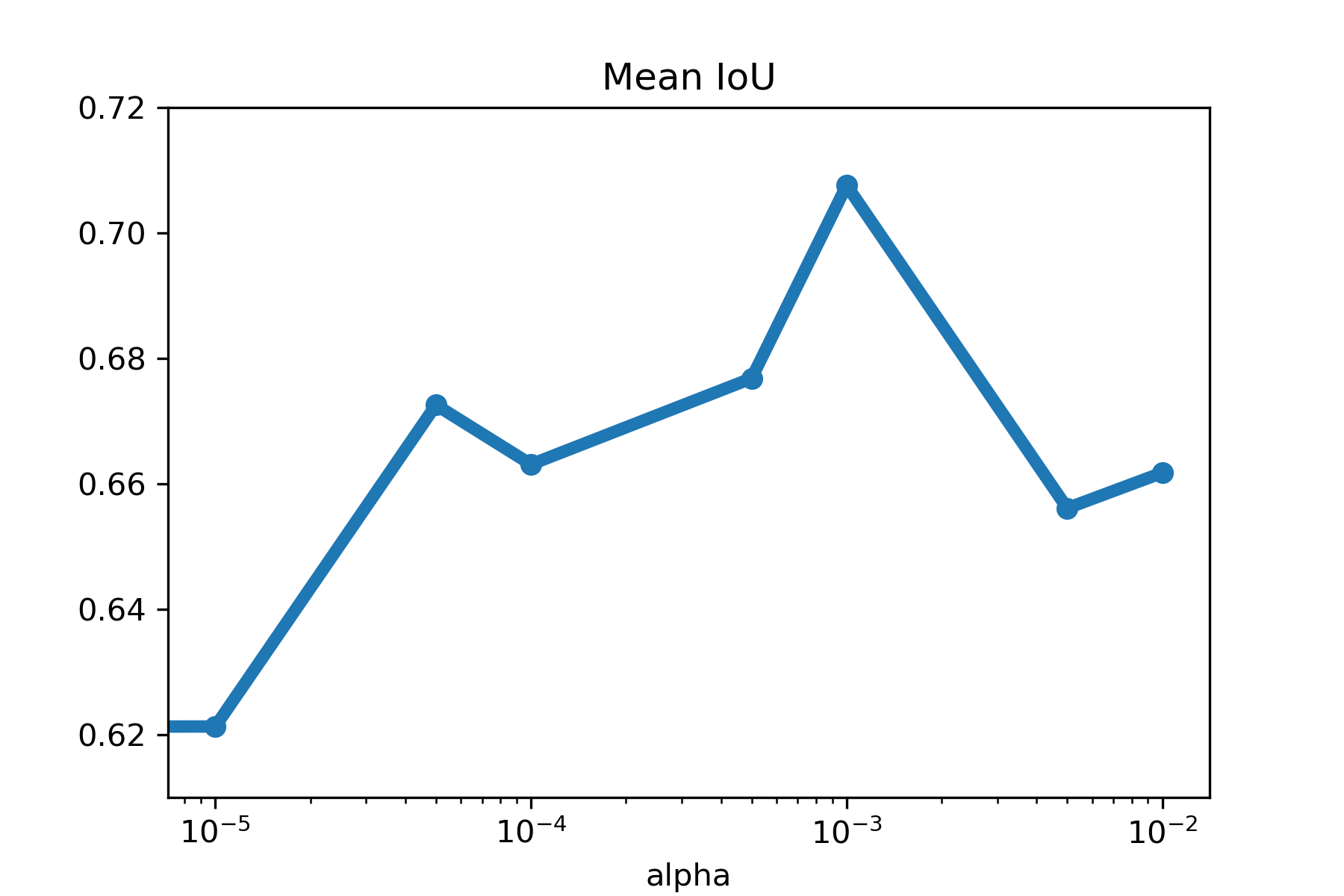}
   \caption{Mean intersection over union of the prediction as a function of the explicit output regularization strength.}
   \label{fig:miou}
 \end{figure}

\begin{figure}[!htb]
   \centering
   \includegraphics[width=0.4\textwidth]{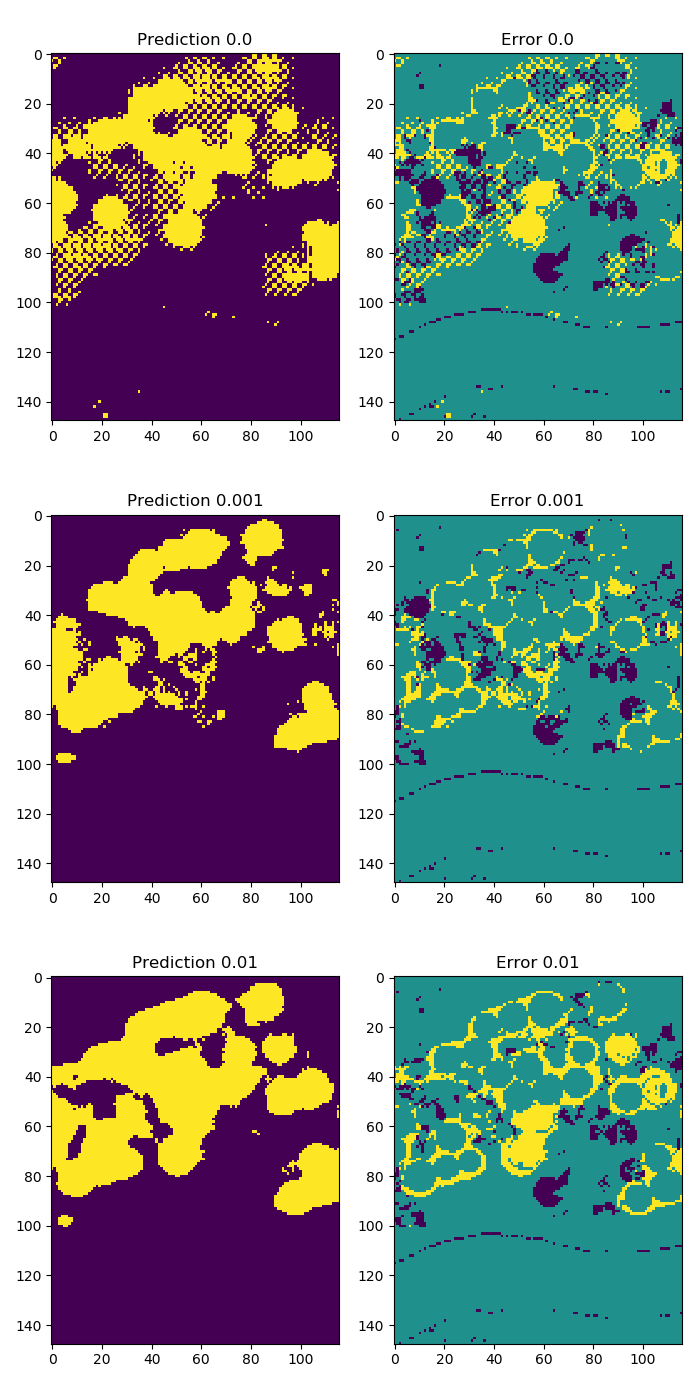}
   \caption{Top: prediction without regularization leads to oscillatory artifacts in the prediction. Middle: results for optimal regularization parameter $\alpha$ in terms of cross-validation for mIoU (Figure \ref{fig:miou}). Bottom: results for very large $\alpha$}
   \label{fig:results}
 \end{figure}

We train a fully reversible network \cite{lensink2019fully} with 10 layers with up to 32 channels for 250 stochastic gradient descent iterations with a decaying learning rate. This is sufficiently many iterations, such that the validation loss is no longer decreasing. Training includes basic data-augmentation via random flips and rotations. We repeat training for a range of $\alpha$  values and assess the results using the mean of the intersection over union per class (mIoU, Figure \ref{fig:miou}). Figure \ref{fig:results} shows results and errors using no regularization, optimal regularization parameter, and a very large $\alpha$. While the best result is not perfect, the simple regularization on the network output yields a significant increase in mIoU.

\section{Conclusions}
We extended the deep connections between partial-differential-equation (PDE) constrained optimization and training neural networks on few data with partial annotation. An insufficient number of annotated pixels leads to oscillatory artifacts in image segmentations. In PDE-constrained optimization, regularization on the model parameters mitigates problems related to insufficient labeling (observations). This carries over directly to the neural network setting as, for example, weight decay. While weight decay regularizes the parameters directly, the final prediction of interest relates indirectly. We showed that explicit regularization on the output of a network serves the same purpose as parameter regularization in PDE-constrained optimization. Brief derivation of the optimization problem and algorithm shows this does not cause computational complications. A hyperspectral example illustrates that simple cross-validation for selecting the regularization strength can improve prediction quality while making minimal assumptions about prior knowledge. Future work should explore more sophisticated regularizers and methods to adapt the regularization strength while training.

{\small
\bibliographystyle{ieee_fullname}
\bibliography{biblio,VideoRefs,HyperSpectralRefs}
}

\end{document}